\documentclass[11pt]{article}

\usepackage{microtype}
\usepackage{graphicx}
\usepackage{subcaption}
\usepackage{booktabs} 

\usepackage[preprint]{acl}

\usepackage{times}
\usepackage{latexsym}
\usepackage[T1]{fontenc}
\usepackage[utf8]{inputenc}
\usepackage{inconsolata}

\usepackage{hyperref}

\usepackage{amsmath}
\usepackage{amssymb}
\usepackage{mathtools}
\usepackage{amsthm}

\usepackage{algorithm}
\usepackage{algorithmic}

\usepackage[capitalize,noabbrev]{cleveref}

\usepackage{placeins}
\usepackage{enumitem}
\theoremstyle{plain}

\theoremstyle{definition}

\theoremstyle{remark}

\newcommand{\sysname}{\textit{DualSQL}}

\usepackage[textsize=tiny]{todonotes}

\usepackage{soul}
\usepackage{xcolor} 
\usepackage[most]{tcolorbox}
\tcbset{breakable}
\usepackage{listings}

\usepackage{ulem}

\definecolor{highlightyellow}{RGB}{255, 250, 205}
\sethlcolor{highlightyellow}

\newtcolorbox{promptbox}[1]{
  enhanced,
  colback=blue!5!gray!10,
  colframe=blue!50!black,
  arc=2mm,
  boxrule=1pt,
  title=#1,
  fonttitle=\bfseries\color{white},
  coltitle=blue!50!black,
  breakable=false,
  width=\textwidth
}

\title{\sysname: Text-to-SQL with Multi-Agent Reinforcement Learning}

\usepackage{latexml}
\iflatexml
\author{
  Shijie Chen\thanks{Work done during Shijie Chen's internship at Google.} \\
  The Ohio State University \\
  \texttt{chen.10216@osu.edu}
  \And
  Yu Gan \\
  Google LLC \\
  \texttt{gany@google.com}
  \And
  Yeounoh Chung \\
  Google LLC
  \And
  Jiani Zhang \\
  Google LLC
  \And
  Quannan Li \\
  Google LLC
  \And
  Sravan Bodapati \\
  Google LLC
  \And
  Cody Greer \\
  Google LLC
  \And
  Yu Su \\
  The Ohio State University
  \And
  Fatma Ozcan \\
  Google LLC \\
  \texttt{fozcan@google.com}
}
\else
\author{
  \textbf{Shijie Chen$^{1}$}\thanks{Work done during Shijie Chen's internship at Google.},
  \textbf{Yu Gan$^{2}$},
  \textbf{Yeounoh Chung$^{2}$},
  \textbf{Jiani Zhang$^{2}$}, \\
  \textbf{Quannan Li$^{2}$},
  \textbf{Sravan Bodapati$^{2}$},
  \textbf{Cody Greer$^{2}$}, \\
  \textbf{Yu Su$^{1}$},
  \textbf{Fatma Ozcan$^{2}$} \\
  $^1$The Ohio State University \qquad $^2$Google LLC \\
  \texttt{chen.10216@osu.edu}, \texttt{\{gany, fozcan\}@google.com}
}
\fi

\begin{document}

\maketitle

\newcommand{\sjc}[1]{{\color{blue}#1}}
\newcommand{\yc}[1]{{\color{purple}#1}}
\newcommand{\ysu}[1]{{\color{cyan}#1}}

\begin{abstract}

State-of-the-art Text-to-SQL systems are typically multi-agent pipelines centered around two fundamental tasks: schema linking and SQL generation.
However, existing work trains separate models for each task, failing to leverage the synergy between these interrelated tasks. 
In this work, we propose \sysname, a new Text-to-SQL system consisting of two agents powered by a single model backbone.
The agents share the same model weights and agentic scaffold, enabling joint optimization through a robust multi-agent reinforcement learning (RL) framework.
We design three database access tools to facilitate effective multi-step reasoning grounded in interactions with the databases.
To improve training and avoid model collapse, we introduce a set of rollout guardrail mechanisms that stabilize multi-agent RL training, enabling \sysname~to keep improving during training.
We also introduce a new SQL correctness metric, robust execution match (REX), to more accurately judge SQL correctness and assign reward signals. 
Trained on only 3755 examples, \sysname-4B achieves an impressive 68.0\% execution accuracy on the BIRD development set, matching previous 7B models. \sysname-8B further improves to 71.1\%, outperforming previous state-of-the-art single-model solutions with 32B parameters. These results demonstrate the strength of joint multi-agent reinforcement learning for building high-performance Text-to-SQL pipelines.
\end{abstract}

\section{Introduction}

Text-to-SQL is a key technology for democratizing access to data analytics for non-expert users through a natural language interface \citep{li2024dawn,luo2025natural}. However, this task is particularly difficult due to several challenges.
First, injecting the entire schema and sampled data from large-scale databases into the context window of language models is prohibitively impractical due to the well-known degradation of reasoning capabilities in long contexts~\cite{du2025context,chung2025evaluating,ling2025longreason} and high prefilling costs. 
Second, SQL queries must be strongly grounded in the database context, requiring precise schema linking for identifying relevant tables and columns, navigating complex schema relationships for formulating correct joins, and a comprehensive understanding of database content for selecting accurate literals~\citep{zhang2025coddllm,su2024tablegpt2}.
Third, the inherent semantic complexity of the Text-to-SQL task complicates the accurate mapping of user intent to valid SQL. This complexity stems from the semantic gap between ambiguous natural language and rigid SQL logic, the requirement for domain knowledge, and ambiguities within the schema itself.

State-of-the-art LLM-based systems address these challenges through a modular design built around schema linking and SQL generation \citep{liu2025xiyan,pourreza2023dinsql,pourreza2025chasesql}.
Within each stage, various methods have been explored, including supervised fine-tuning \citep{li2025omnisql}, in-context learning \citep{talaei2024chess,pourreza2025chasesql}, and reinforcement learning from execution feedback \citep{pourreza2025reasoningsql,yao2025arctictext2sqlr1}.
However, these methods optimize the two stages in isolation, pointing to a critical gap in joint optimization of the end-to-end pipeline.

We argue that, despite differing task formulations, schema linking and SQL generation share two core capabilities: semantic retrieval over complex relational schemas and reasoning grounded in database context.
Therefore, instead of building separate models, it is better to treat these tasks as dual facets of a unified agent that can operate on databases, learn shared knowledge representations, and transfer reasoning skills.

In this work, we propose \sysname, a Text-to-SQL framework that consolidates schema linking and SQL generation into a unified agentic pipeline powered by a single, parameter-shared open-weight LLM.
Inspired by the iterative workflow of human database developers, we equip \sysname~with three database access tools: metadata profiling, full-text search, and SQL execution. Rather than relying on rigid, pre-defined workflows, \sysname~learns to autonomously interact with databases, dynamically verifying and correcting its reasoning through tool-use feedback. 

The entire system is jointly trained with a multi-agent reinforcement learning framework, supported by an asynchronous rollout system that decouples synchronization between stages. We find that standard stabilization techniques such as importance sampling correction~\cite{mathematical_verl_2025,liu-li-2025-rl-collapse,zheng2025stabilizing} are insufficient to prevent structural degeneration in our multi-agent, multi-turn tool-use settings. We therefore introduce a set of agentic rollout guardrails, including strict format checking with rollout cutoff and error-focused loss masking. Furthermore, to provide cleaner reward signals than the standard execution accuracy (EX), which is overly sensitive to benign formatting variations and insensitive to duplicates, we introduce Robust Execution Match (REX). REX can be computed deterministically to efficiently assess exact execution equivalence, preventing the reinforcement of false positives during training.

By combining this stable agentic RL framework with the unified multi-agent formulation, \sysname~demonstrates that open-weight models can achieve strong Text-to-SQL capabilities without relying on massive parameter counts or multi-model ensembles. On the challenging BIRD benchmark, \sysname-4B achieves a competitive 68.0\% EX, while \sysname-8B establishes a new state-of-the-art for single-model solutions at 71.1\% EX, rivaling systems that utilize 32B parameters.

To summarize, our main contributions are:
\begin{itemize}[noitemsep,nolistsep]
    \item We introduce a multi-agent Text-to-SQL framework driven by a shared LLM backbone and three database access tools.
    \item We introduce comprehensive agentic rollout guardrails that successfully stabilize multi-agent RL training, and Robust Execution Match, which provides more accurate SQL evaluation.
    \item We demonstrate that joint multi-agent optimization significantly extends the performance ceiling of single-model pipelines, achieving highly competitive performance with small model sizes.
\end{itemize}


\section{Related Work}
\subsection{LLM-based Text-to-SQL}

Recent advancements in Text-to-SQL have progressed from simple prompt engineering on general-purpose LLMs to specialized multi-agent systems~\citep{biswal2026agentsm,wang2025agentar,deng2025reforce}.
Initial efforts utilized in-context learning (ICL) to optimize the zero-shot and few-shot performance of proprietary LLMs like GPT-4~\citep{gao2023dailsql, pourreza2023dinsql,chung2025islongcontext}.
Subsequent context engineering strategies~\citep{chung2025islongcontext,askdata2025} highlighted the criticality of schema linking via augmented metadata for accurate SQL generation. 
Unlike ICL-based methods, which rely on passive context engineering, \sysname~learns to perform iterative schema linking with tool use.
To handle query complexity, the field shifted toward multi-agent collaboration~\citep{wang2025mac, pourreza2025chasesql} for task decomposition and external tool leverage.
To reduce API dependency and improve cost-efficiency, supervised fine-tuned approaches emerged~\citep{ma2025db}, utilizing large-scale synthetic datasets to align open-source models with the Text-to-SQL task~\citep{li2024codes,li2025omnisql}. 
This paradigm was further refined by integrating RL with execution feedback to optimize reasoning chains and bridge the performance gap with larger models~\citep{ma2025sqlr1, yao2025arctictext2sqlr1, zhang2025rewardsql}.
While the latest solutions synthesize agentic decomposition and RL~\citep{liuskyrl,xu2025mtirsql,yang2025marssql}, they optimize agents for different stages separately, neglecting their synergy.

\sysname~differs from existing work by (1) introducing the robust execution match metric for precise SQL correctness evaluation, (2) providing a comprehensive database manipulation toolset, and (3) establishing a robust multi-agent RL framework for joint optimization of the Text-to-SQL pipeline.

\subsection{Reinforcement Learning with Verifiable Rewards}
Reinforcement Learning with Verifiable Rewards (RLVR) has become a cornerstone for enhancing LLM reasoning in domains with verifiable outcomes, such as mathematics~\citep{shao2024deepseekmath} and coding~\citep{guo2025deepseekr1}.
To address inherent training instabilities, recent research has introduced optimization techniques ranging from decoupled clipping and dynamic sampling~\citep{yu2025dapo} to variance-based rollout down-sampling~\citep{xu2025not}.
In multi-turn agentic scenarios, methods such as entropy-based adaptive branching~\citep{dong2025agentic} and void-turn filtering~\citep{xue2025simpletir} have been proposed to mitigate distribution shifts, while other studies have investigated the role of high-entropy tokens~\citep{wang2025beyond}, the boundaries of reasoning capabilities versus sampling efficiency, and curriculum learning for autonomous agents~\citep{nguyen2025sfr, zheng2025parallel}. 
The complex and noisy database environments pose new challenges to agentic RL stability for text-to-SQL systems. Our work contributes an efficient rollout implementation and a set of agentic rollout guardrails, which effectively enhance the efficiency and stability of multi-agent RL, leading to better performance via prolonged exploration.



\begin{figure*}
  \centering
  \includegraphics[width=0.85\linewidth]{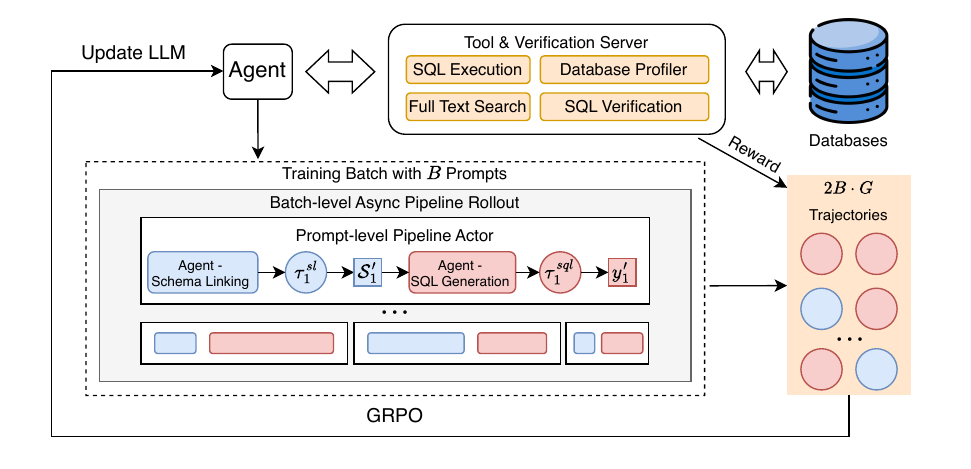}
  \caption{The multi-agent training framework for \sysname. $sl$ and $sql$ denote schema linking and SQL generation tasks, respectively. The two agents are driven by the same LLM and tools.}
\end{figure*}

\section{\sysname~Agentic System}

This section introduces the agentic system of \sysname.
We introduce a unified agentic formulation for the sub-agents in Text-to-SQL systems: schema linker and SQL generator.
Each agent is powered by the same backbone LLM that is equipped with a set of database access tools, engaging in multi-turn interactions to understand database context and reason about the question.


\subsection{Database Access Tools}
\label{sec:tools}
Inspired by the iterative workflow of human database developers, we equip the agents with three database access tools. 
Implementation details, including JSON serialization of tool outputs and asynchronous tool hosting, are in Appendix~\ref{appendix:tool_impl}.

\textbf{SQL Executor.} The primary interface between the agents and the database. The agents use it for (i) execution feedback to verify syntactic and semantic correctness, (ii) entity grounding to inspect literal representations, and (iii) ad-hoc schema exploration via \texttt{sqlite\_master} or \texttt{PRAGMA}. A content-aware truncation policy keeps standard retrievals within context limits while preserving full schema-query responses.

\textbf{Full Text Search.} To bridge ambiguous or abbreviated mentions in user queries to literal database content, this tool performs fuzzy matching over schema names and cell values using SQLite FTS5 inverted indices. It returns the matched schema elements and up to five matching values per column.

\textbf{Database Profiler.} This tool retrieves offline-prepared database metadata. For each table and column, we use an LLM to produce a description summarizing semantic intent and to infer foreign-key relationships that may be missing from the DDL. We additionally pre-compute column-level statistics (distinct counts, types, null ratios). Offloading this analysis offline frees the agent to focus on query logic at runtime.

\subsection{\sysname~Agents}
\sysname~features a multi-agent pipeline consisting of two agents, both of which are powered by the same backbone LLM and can engage in multi-turn interactions with the database via the tools introduced in Section~\ref{sec:tools}:

\textbf{Schema linker:} Given the natural language question $Q$ and the complete database schema $\mathcal{S}=\{\mathcal{T}, \mathcal{C}\}$, where $\mathcal{T}$ and $\mathcal{C}$ are the sets of tables and columns, the schema linker identifies the relevant database schema $\mathcal{S}'\subseteq \mathcal{S}$ to reduce the distraction in large databases and ensures the subsequent SQL generator operates within a focused context window.

\textbf{SQL generator:} The SQL generator generates a SQL query $y'$ based on the question $Q$ and the linked schema $\mathcal{S}'$. By modeling schema linking and SQL generation as a joint stochastic process, we encourage the generated SQL to be grounded in the correct schema elements, reducing hallucination from irrelevant table schemas.


Despite its simplicity, this unified design proves effective and allows us to jointly optimize the entire pipeline with our multi-agent RL framework.

\section{Multi-Agent RL Framework for Optimizing Text-to-SQL Pipelines}

The unified agent design enables joint optimization of the two agents by training a single LLM.
In this section, we introduce our multi-agent RL recipe, including an asynchronous rollout system, rollout guardrails, reward design, and training objectives.

\subsection{Asynchronous Rollout System}
\label{sec:async_rollout}

Considering the dependency structure of the above Text-to-SQL pipeline and that agent trajectories vary greatly in length, we design an asynchronous rollout system which manages asynchronous rollout for parsing pipelines. We treat schema linking and SQL generation as an integral stochastic process, which avoids synchronization between stages and improves throughput of multi-agent rollout.

For each batch of $B$ prompts, the \textit{batch-level controller} manages asynchronous rollout of different prompts, each supported by the prompt-level pipeline actor.
The \textit{prompt-level pipeline actor} executes individual Text-to-SQL pipelines. It flexibly supports rollout of a single task or a pipeline chaining schema linking and SQL generation.
With a group size of $G$, we sample a total of $2B \cdot G$ trajectories in each optimization step.
These trajectories are later grouped by task and used for joint optimization of the policy model.


\subsection{Agentic Rollout Guardrails}
In our preliminary experiments, we observe that multi-agent RL training is much more unstable than non-agentic or single-agent RL training, shown by frequent collapse in trajectory format in one or both tasks. Existing work advocates for variants of sequence-level importance sampling or negative sampling to calibrate the distribution mismatch between rollout and policy models, aiming to stabilize training~\cite{liu-li-2025-rl-collapse, mathematical_verl_2025}. However, we find that representative collapse patterns, such as the incorrect use of format tokens, endless loops, and incorrect tool calls, persist even with these calibration techniques.
To improve the robustness of multi-agent RL training, we design a set of \textbf{agentic rollout guardrails}, including strict format checking, rollout cutoff, and loss masking. 
%

\textbf{Strict Format Checking.} To maintain format correctness of agentic rollouts, we design a strict format checking mechanism that enforces the following requirements:
\begin{itemize}[noitemsep]
  \item In each turn, the agent should generate a non-empty \texttt{<think></think>} block for effective reasoning.
  \item In each turn, the agent should generate either a \texttt{<tool\_call></tool\_call>} block for tool calls or a single \verb|```...```| code block for returning final answers, such as a linked schema or a SQL query.
  \item If a tool call is generated, it should be formatted as a JSON string following the pre-defined tool call format, including a valid tool name and tool arguments. Moreover, the tools are idempotent and the same tool should not be called more than once with the exact same argument.
  \item If a code block is generated, the content after the code block should not constitute more than half of the total length in that turn. This requirement prevents the model from collapsing into generating redundant or even endless explanations after the solution is given.
\end{itemize}

If a trajectory violates any of the above constraints, we set the format correctness reward $R_f$ to 0. Otherwise, $R_f=1$. In this way, format checking helps reduce degeneration and mode collapse during training.

\textbf{Rollout Cutoff.} Allowing rollout to proceed after format violations can lead to a reward hacking pattern, where agents learn that format errors are acceptable as long as the final answer is correct.
Moreover, the malformed actions pollute reward signals, causing the policy model to reinforce invalid actions and eventually reach an unrecoverable state.
To prevent this, we stop rollout immediately when detecting any format constraint violation, which acts as a hard filter on the exploration space, ensuring that the agent is only rewarded when strictly adhering to the format requirements.

\textbf{Loss Masking.} Penalizing an entire multi-turn trajectory for a terminal format error introduces noise in reward signals, as it may discourage the valid reasoning steps that preceded the error.
At the same time, rewarding correct trajectories with intermediate format or tool call errors also risks reinforcing bad behaviors.
Therefore, we mask out the loss for previous turns if a trajectory ends with a turn with a format error to concentrate learning signals on format error turns.
This approach effectively teaches the agent to unlearn malformed actions before improving reasoning correctness.


At test time, we add meaningful error messages as tool responses to the trajectory upon occasional errors, such as JSON parsing errors or argument type mismatches.
In this way, the agent can correct its tool-use mistakes and continue reasoning.


\subsection{Reward Design}
\textbf{Schema Linking.} We design a reward function that encourages high recall.
Given the ground truth schema items $S^{*}$ and the linked schema items $S'$, we compute the schema linking reward as:

{\small
\begin{equation}
  R_{sl} =
    \begin{cases}
      1 & \text{if $S^{*}$ = $S'$}\\
      0.9 - \max(0.2-\frac{|S^{*} \cap S'|}{|S'|}, 0) & \text{if $S^{*} \subseteq S'$}\\
      0.7 - \max(0.2-\frac{|S^{*} \cap S'|}{|S'|}, 0) & \text{if $T^{*} \subseteq T'$}\\
      0.5 \cdot \frac{2|S^{*} \cap S'|}{|S^{*}|+|S'|} & \text{otherwise}
    \end{cases},
\end{equation}
}
where $T^{*}$ and $T'$ are the sets of ground truth and linked tables, respectively.

This reward function gives full credit to perfect matches. Then, partial credit is given for complete schema-level and table-level recall with a penalty for low precision. Finally, we give partial credit based on F1 score.

\textbf{SQL Generation.} We use the correctness of the predicted SQL query $y'$ as the reward signal for SQL generation. To avoid degeneration into single-turn reasoning, we add a bonus for correct trajectories that use tools $R_{tool}=1$.

\begin{equation}
  R_{sql} =
    \begin{cases}
      1 +0.1\cdot R_{tool}& \text{if $Eq(y'$,$y^{*})$}\\
      0 & \text{otherwise}
    \end{cases}       
\end{equation}
where $y^{*}$ is the ground truth SQL query and $Eq(\cdot)$ indicates whether two SQL queries are equivalent.

The execution match (EX) metric, widely used in Text-to-SQL benchmarks like Spider and BIRD, first converts SQL query result sets into a set of literals and evaluates equivalence.
This process overlooks the important discrepancies brought by duplicated rows and row ordering, leading to false positive evaluations.
Moreover, EX also does not tolerate column order permutations, which introduces false negative signals.

We propose \textbf{robust execution match} (REX), a stricter metric based on data similarity, which evaluates equivalence by finding the maximum matching between the result sets of two SQL queries.
If the question does not demand row order matching, REX first sorts the result sets by high-cardinality columns.
Then, REX tries to find a permutation of columns that leads to the maximum matching.
If a perfect matching is found, the two result sets are considered equivalent.
We decide whether a question requires row ordering matching based on the presence of the \texttt{ORDER BY} keyword in ground truth SQL queries.
Algorithm details are available in Appendix~\ref{appendix:data_matching_algorithm}.


\textbf{Format Reward and Length Penalty.}
We use the soft overlong punishment introduced by DAPO to penalize overlong trajectories. Additionally, the length penalty also implicitly incentivizes the agent to avoid unnecessary self-correction steps by assigning higher marginal rewards to concise, error-free trajectories.

{\scriptsize
\begin{equation}
  R_{l} =
  \begin{cases}
        0 & |\tau| \leq L_{max} - L_{cache}\\
        \frac{(L_{max} - L_{cache}-|\tau|)}{L_{cache}} & L_{max} - L_{cache} < |\tau| \leq L_{max}\\
        -1, & |\tau| > L_{max}
  \end{cases} 
\end{equation}
}

We only apply the length penalty to trajectories with the correct format to avoid false positive signals. The final reward is computed as:
\begin{equation}
  R = \max(R_{task}, 0.1 \cdot R_f) + \lambda_l R_l \cdot R_f,
\end{equation}
where $R_{task}\in\{R_{sl}, R_{sql}\}$ is the task-specific reward. $\lambda_l=0.2$ is the weight for the length penalty. We further clip negative rewards to 0 to prevent undesirable trajectories from having a positive advantage.

\subsection{Training Objectives}
We optimize \sysname~with GRPO~\citep{shao2024deepseekmath}, incorporating token-mean loss and clip-higher~\citep{yu2025dapo} into the training objective, and sequence-level masked importance sampling~\citep{liu-li-2025-rl-collapse} for enhancing training stability. For each prompt $x$, we sample $2G$ trajectories from $\pi_\theta$ ($G$ per task $t\in\{sl, sql\}$) and compute a task-specific group-relative advantage $\hat{A}^{(i)}_t$ for policy updates.

\section{Experiments}

\subsection{Experiment Setup}
\textbf{Training Data Preparation}
We train our models on the officially revised training set of BIRD. 
We filter out 2715 trivial examples where Qwen3-8B has $Pass^{16}=1$ and 466 intractable examples where Gemini-2.5-Pro has $\text{Pass@32}=0$ when evaluated in both single-turn and agentic Text-to-SQL settings.
This leads to a training set of 3376 examples.
We further manually analyze the intractable questions and correct 384 annotation errors as augmented training data, leading to 3755 training examples after deduplication.

\textbf{RL Setup}
We implement our multi-agent RL framework based on VeRL \citep{sheng2024hybridflow}.
We use a learning rate of 2e-6, linear warmup of 20 steps, batch size of 256, and group size of 16 during training.
For rollout, we set the temperature to 1.0, $\text{top-k}=20$ and $\text{top-p}=0.99$ to encourage exploration.
We use Qwen3 \citep{qwen3} 4B and 8B models as the base models for training.
The trajectory length limit is set to 32K tokens and the overlong buffer size $L_{cache}$ to 4K tokens.
More training details, including the training performance curve, can be found in Appendix~\ref{appendix:training_details}.

\subsection{Experiment Results}

We follow the same schema serialization procedure as OmniSQL~\citep{li2025omnisql} to be consistent with previous work and use OmniSQL's prompt when evaluating base LLMs. By default, we report the average performance over 8 runs.

\subsubsection{In-Domain Evaluation}

\textbf{Text-to-SQL} We compare $\sysname$ with existing RL-based single-model Text-to-SQL methods on the BIRD development set (BIRD-Dev), including single-turn models like SQL-R1 \cite{ma2025sqlr1}, Reasoning-SQL \cite{pourreza2025reasoningsql} and Arctic-Text2SQL-R1 \citep{yao2025arctictext2sqlr1}, and the single-agent MTIR-SQL \citep{xu2025mtirsql} with only the SQL execution tool.

\begin{table}[t]
  \small
  \centering
  \begin{tabular}{ccccc}
  \toprule
  Method & Size & Agent & EX\\
  \midrule
  MTIR-SQL & 4B & $sql$ & 64.4 \\
  Reasoning-SQL & 7B & $\times$ & 64.0 \\
  SQL-R1 & 7B & $\times$ & 66.6 \\
  Arctic-Text2SQL-R1 & 7B & $\times$ & 68.9 \\
  MTIR-SQL & 8B & $sql$ & 64.6 \\
  Reasoning-SQL & 14B & $\times$ & 65.3 \\
  SQL-R1 & 14B & $\times$ & 67.1 \\
  MTIR-SQL & 14B & $sql$ & 68.1 \\
  Arctic-Text2SQL-R1 & 14B & $\times$ & 70.1 \\
  Arctic-Text2SQL-R1 & 32B & $\times$ & 70.5 \\
  \midrule
  $\sysname$ (Ours) & 4B & $sl+sql$ & 68.0 \\
  $\sysname$ (Ours) & 8B & $sl + sql$ & \textbf{71.1} \\
  \bottomrule
  \end{tabular}
  \caption{\label{tab:nl2sql} Performance on the BIRD development set.}
\end{table}


As shown by Table~\ref{tab:nl2sql}, with multi-agent RL training of both schema linking and SQL generation, $\sysname$ achieves impressive Text-to-SQL performance with small model sizes.
$\sysname$-4B reaches 68.0\% EX, matching prior 7B baselines.
$\sysname$-8B further improves performance to 71.1\% EX, surpassing previous methods built on 14B or 32B models and establishing a new single-model state-of-the-art.
Breaking down by question difficulty (Appendix~\ref{appendix:difficulty_breakdown}), our multi-agent formulation consistently improves over the SQL-only $\sysname_{sql}$ baseline across all difficulty levels, with the largest gain on challenging questions (3.0\% EX for the 8B model).

\begin{table}[t]
  \small
  \centering
  \begin{tabular}{cccccc}
  \toprule
  Method &  Precision & Recall & F1 & C-Recall\\
  \midrule
  $\sysname_{sl}$-8B &  91.7 & 89.9 & 90.0 & 64.7 \\
  \midrule
  $\sysname$-4B & 92.6 &  89.5 & 90.3 & 63.8 \\
  $\sysname$-8B & \textbf{93.1} &  \textbf{90.0} & \textbf{90.8} & \textbf{65.8} \\
  \bottomrule
  \end{tabular}
  \caption{\label{tab:schema_linking} Schema linking performance on BIRD-Dev. C-Recall denotes complete recall.}
\end{table}

\textbf{Schema Linking} 
We evaluate the schema linking performance of \sysname~on the BIRD development set by comparing linked schema $\mathcal{S'}$ with the schema used by the ground truth query $\mathcal{S}^{*}$.
In addition to precision, recall, and F1 score, we also report complete recall rate (C-Recall), which measures the percentage of examples where all ground truth schema entities are retrieved by the agent.

Table \ref{tab:schema_linking} shows the schema linking performance of \sysname.
We compare the performance of standalone RL for the schema linking agent ($\sysname_{sl}$-8B) and multi-agent RL.
As shown by the table, \sysname-8B can accurately find relevant schema items, achieving over 90\% F1 score in schema linking.
Multi-agent RL training additionally boosts precision by 1.4\% and C-Recall by 1.1\%, demonstrating the synergy between agents.

\begin{table}[t]
  \small
  \centering
  \setlength{\tabcolsep}{2pt}
  \begin{tabular}{lcc}
  \toprule
  Model & EX & REX\\
  \midrule
  $\sysname$-4B & $68.0_{\pm 0.74}$ & $64.0_{\pm 0.75}$ \\
  $\sysname$-4B w/ oracle schema & $71.7_{\pm 0.40}$ & $66.8_{\pm 0.64}$ \\
  $\sysname_{sql}$-8B & $69.6_{\pm 0.28}$ & $66.7_{\pm 0.30}$ \\
  $\sysname_{sql}$-8B w/ oracle schema & $75.7_{\pm 0.64}$ & $71.1_{\pm 0.60}$ \\
  $\sysname$-8B & $71.1_{\pm 0.23}$ & $68.6_{\pm 0.21}$ \\
  $\sysname$-8B w/ oracle schema & $\textbf{76.5}_{\pm 0.50}$ & $\textbf{72.1}_{\pm 0.48}$ \\
  \bottomrule
  \end{tabular}
  \caption{\label{tab:oracle_schema} SQL generation performance on BIRD-Dev when paired with an oracle schema linker.}
\end{table}

\textbf{Linking Errors and Parsing Errors.}
We conduct a manual error analysis on \sysname-8B's errors and find that only 15.3\% of them are due to schema linking errors such as wrong columns and tables, despite C-Recall being only 65.8\%.
This discrepancy exists because C-Recall is a strict and over-pessimistic metric. In practice, we find \sysname~can use tools to dynamically recover missing schema context during its multi-turn reasoning process in the SQL generation stage, compensating for imperfect schema linking results.

To quantify the impact of imperfect schema linking, we evaluate \sysname-8B's performance with oracle schema linkers (Table~\ref{tab:oracle_schema}). \sysname-8B's performance can be improved by 5.4 points, demonstrating sizable headroom for improvement in schema linking.
We also notice that \sysname~maintains its advantage over $\sysname_{sql}$ by 0.8 points in EX, indicating that joint multi-agent RL indeed improves reasoning capabilities.

\subsubsection{Out-of-domain Evaluation}


\begin{table}[ht]
\centering
\footnotesize
\setlength{\tabcolsep}{1.5pt}
\begin{tabular}{lcccc}
  \toprule
  & \multicolumn{2}{c}{BIRD-Dev} & \multicolumn{2}{c}{Spider} \\
  Model & EX & REX & EX & REX \\
  \midrule
  Qwen3-4B &  $59.6_{\pm 0.49}$ & $52.8_{\pm 0.52}$ & $78.9_{\pm 0.22}$ & $81.4_{\pm 0.25}$ \\
  \sysname-4B & $\textbf{68.0}_{\pm 0.74}$ & $\textbf{64.0}_{\pm 0.75}$ & $\textbf{81.9}_{\pm 0.30}$ & $\textbf{85.3}_{\pm 0.27}$ \\
  \midrule
  Qwen3-8B & $60.9_{\pm 0.60}$ & $56.6_{\pm 0.47}$ & $78.9_{\pm 0.33}$ & $76.1_{\pm 0.35}$ \\
  Single-turn RL & $67.3_{\pm 0.46}$ & $61.3_{\pm 0.39}$ & $82.3_{\pm 0.37}$ & $78.7_{\pm 0.35}$ \\
  ${\sysname}_{sql}$-8B & $69.6_{\pm 0.28}$ & $66.7_{\pm 0.30}$ & $82.7_{\pm 0.21}$ & $80.6_{\pm 0.31}$ \\
  $\sysname$-8B & $\textbf{71.1}_{\pm 0.23}$ & $\textbf{68.6}_{\pm 0.21}$ & $\textbf{83.1}_{\pm 0.29}$ & $\textbf{80.8}_{\pm 0.18}$ \\
  \bottomrule
\end{tabular}
\caption{Comparing different RL training settings on BIRD-Dev and Spider.}
\label{tab:ood_benchmark}
\end{table}

Our models are trained on data from only the BIRD training set. To evaluate the generalization ability of our models, we also test \sysname~on Spider's test set \citep{yu2018spider} and compare single-turn RL, single-agent RL, and multi-agent RL when using the same training data and parameter setup.
As shown by Table \ref{tab:ood_benchmark}, we observe a significant advantage of single-agent RL over single-turn RL on the 8B models.
In addition, our joint multi-agent RL training method further improves the text-to-SQL performance on both benchmarks, demonstrating that the tool-use and reasoning capabilities learned during training can generalize to Spider.

\subsubsection{Ablation Studies}

\textbf{Effectiveness of Multi-Agent RL:}
To understand the effectiveness of our multi-agent RL framework, we compare RL outcomes using Qwen3-8B under 3 settings: (1) single-turn Text-to-SQL with reasoning, (2) multi-turn agentic Text-to-SQL, and (3) multi-agent Text-to-SQL (schema linking + SQL generation) trained with the same reward function and hyperparameter settings.

Using the same training data and reward signal, we observe a significant performance gain of 2.3\% EX and 5.4\% REX when moving from single-turn Text-to-SQL to multi-turn agentic Text-to-SQL, demonstrating the effectiveness of our agentic Text-to-SQL framework. 
Multi-agent RL training of both schema linking and SQL generation further improves the performance by 1.5\% EX and 1.9\% REX on BIRD-Dev, showing the benefit of joint optimization of multiple agents.

\textbf{Effectiveness of Robust Execution Match:} 
To study the effectiveness of the proposed robust execution match as a reward signal, we compare RL outcomes in the single-agent setting when using EX or REX as $Eq(\cdot)$ in the reward function.
\begin{table}[ht]
  \small
  \centering
  \begin{tabular}{ccc}
  \toprule
  $Eq(\cdot)$ & EX & REX\\
  \midrule
  EX & 69.2 & 63.7 \\
  REX & \textbf{69.6} & \textbf{66.7} \\
  \bottomrule
  \end{tabular}
  \caption{\label{tab:ablation_EX_REX} Comparison of performance on BIRD-Dev when using EX or REX as $Eq(\cdot)$ in the reward function for single-agent Text-to-SQL RL. We test on 8B models.}
\end{table}

As shown by Table \ref{tab:ablation_EX_REX}, using the high-fidelity REX metric as the reward signal leads to better RL outcomes measured in both EX and REX metrics.
Through error analysis, we find that the model trained with EX as the reward signal struggles with questions where using \texttt{DISTINCT} and \texttt{ORDER BY} clauses is required, which are also the scenarios where EX fails to distinguish.
This demonstrates that a cleaner reward signal consistently leads to better training outcomes, although the target benchmark is measured using a coarser metric.

\textbf{Effectiveness of Agentic Rollout Guardrails:}
To test the effectiveness of the proposed agentic rollout guardrails, we compare RL training runs with and without the guardrails and report execution accuracy on BIRD-Dev during training.

\begin{figure}[ht]
  \centering
  \includegraphics[width=\linewidth]{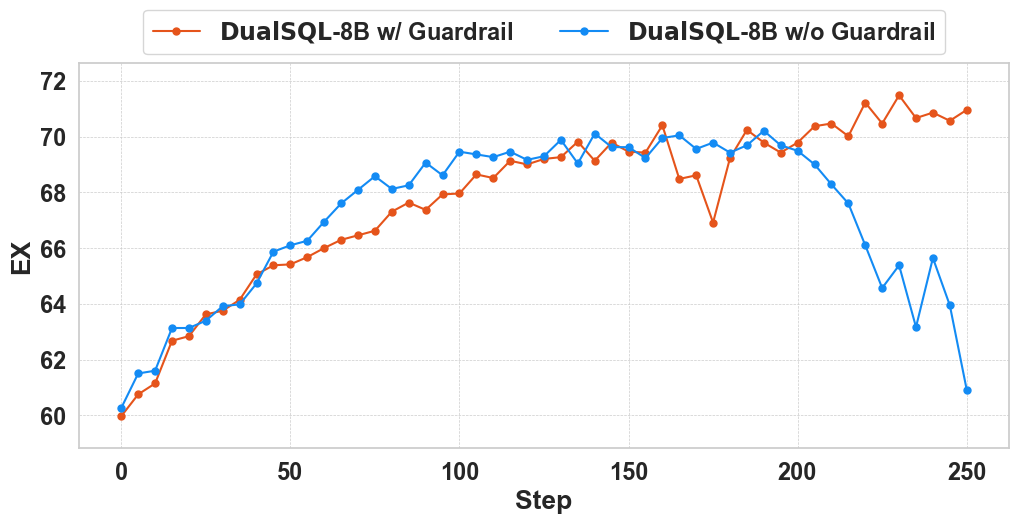}
  \caption{\label{fig:ablation_guardrails} Ablation study on the effectiveness of agentic rollout guardrails. We report execution accuracy (EX) on BIRD-Dev during training. Sequence-level masked importance sampling is enabled for both models.}
\end{figure}

Fig.~\ref{fig:ablation_guardrails} shows that agentic rollout guardrails prevent mode collapse before convergence by guiding the model to unlearn erroneous behavior immediately at the error turns. This allows the agent to maintain format correctness and reach higher performance via sustained training. In contrast, malformed actions are retained in trajectories with high rewards without the guardrails, which are reinforced during training and eventually lead to behavioral collapse.

\subsubsection{\sysname~Reasoning Strategies}

To characterize the behavior of \sysname, we analyze the reasoning trajectories and tool usage patterns that emerged through joint multi-agent reinforcement learning.
Table~\ref{tab:tool_use_distribution} summarizes the average number of tool calls made by \sysname~in each stage on BIRD-Dev. An ablation study of individual tools is available in Appendix~\ref{appendix:ablation_tools}.

\begin{table}[ht]
  \small
  \centering
\begin{tabular}{cccccc}
\toprule
Stage & SE & FTS & DP &Total \\
\midrule
Schema Linking & 0.25 & 0.00 & 1.00 & 1.26 \\
SQL Generation & 1.22 & 0.05 & 0.49 & 1.76 \\
\bottomrule
\end{tabular}
\caption{\label{tab:tool_use_distribution}Average number of tools used by \text{\sysname-8B} on BIRD-Dev. SE: SQL Execution, FTS: Full Text Search, DP: Database Profiler.}
\end{table}

We highlight two primary strategies that distinguish \sysname~from single-turn baselines. (Please see Appendix~\ref{appendix:example_trajs} for example agent trajectories.)

\textbf{Interactive Schema Linking:} Rather than operating on a potentially hallucinated mental model derived from static DDL, the agents employ an interactive schema linking strategy. By leveraging the Database Profiler to inspect table metadata and the SQL Executor to sample specific rows, the agents actively resolve lexical mismatches and verify columns, tables, and join paths before generating the SQL code. The strategic separation between schema linking and SQL generation reduces the attention dilution of relevant information within massive input sequences in the SQL generator, allowing it to operate within a pruned, high-relevance context window. Additionally, our two-stage approach achieves robust failure recovery by augmenting self-correction with cross-validation, mitigating the risks of relying solely on self-correction~\cite{zhang-etal-2025-understanding}. Should the pre-filtered schema from the schema linker prove insufficient during execution, the SQL generator retains the capability of using tools to refine its schema understanding.

\textbf{Execution-guided Error Correction:} Unlike the standard single-turn reasoning models that are prone to syntactically valid queries returning incorrect results, \sysname~emulates the behavior of human developers to transform SQL generation from a static parsing task into a dynamic hypothesis-testing loop. By analyzing intermediate execution feedback, the agents detect discrepancies (e.g., numerical data stored as strings, non-standard date formats, or string values in a different language) and iteratively refine their queries with precise dialect-specific adjustments, such as type casting or string manipulation.
It ensures the final SQL is consistent with users' intents and the contents stored in the database.

\section{Conclusion}
We propose \sysname, a multi-agent Text-to-SQL system optimized by a robust reinforcement learning framework.
\sysname~unifies schema linking and SQL generation as two agents powered by a single LLM backbone and a set of database access tools, enabling effective reasoning over databases and joint optimization through RL.
\sysname-4B reaches 68.0\% EX on BIRD-Dev, matching prior 7B baselines, while \sysname-8B sets a new single-model state-of-the-art of 71.1\% EX, surpassing methods built on 32B models. These results demonstrate that joint multi-agent optimization is an effective path toward stronger Text-to-SQL systems.

We see three promising directions for future work: (1) better reward design for multi-agent reinforcement learning, including fine-grained process reward signals and inter-agent credit assignment mechanisms; (2) scaling training data via synthesis of large-scale databases and challenging Text-to-SQL questions; and (3) more effective multi-agent RL recipes for training larger models and supporting longer-horizon trajectories.

\section*{Limitations}

\textbf{Model backbone.} We instantiate \sysname~on the Qwen3 family because (1) it natively supports both extended reasoning and tool calling, two key capabilities our agentic formulation requires; and (2) it is well supported by open-source training frameworks like VeRL. Nevertheless, we believe our training recipe to be applicable to any backbone LLMs with these capabilities, including open-weight and proprietary ones.

\textbf{Reward design.} REX substantially improves over EX as a reward signal, as confirmed by our ablations. However, it is still a proxy for SQL equivalence checking and could produce false signals.  In addition, as with any outcome-based reward function, our reward design lacks fine-grained process-level reward signals which could be important for training stronger Text-to-SQL agents in the future.

\textbf{Security.} We restrict the agents to read-only database access throughout training and evaluation. As with any LLM-driven system that interfaces with live data sources, deploying \sysname~in production warrants standard precautions against query-induced resource exhaustion and adversarial prompt injection.

\section*{Artifact and Data Usage}
This work leverages datasets and code from previous research, including the BIRD Dataset \citep{li2023bird} (CC BY-SA 4.0), the Spider dataset \citep{yu2018spider} (CC BY-SA 4.0), code from the VeRL (Apache 2.0) framework \citep{sheng2024hybridflow}, and the Qwen3 (Apache 2.0) family of models \citep{qwen3}. These data, software, and models have permissive licenses for use in academic research.

\section*{AI Assistant Usage}

The authors used AI assistants for editing and improving the language of the paper. All presented content is reviewed by the authors.

\bibliography{references}

\newpage
\appendix
\section{More Tool Implementation Details}
\label{appendix:tool_impl}

To optimize the model's consumption of tool results, we serialize all tool results using the \texttt{JSON} format, minimizing the distribution gap between tool results and the model's pre-training data. We further host the tools as an independent asynchronous web service, isolating the high latency of database operations from the GPU-intensive LLM inference process so that database access does not bottleneck GPU utilization during rollout.

\textbf{SQL Executor.} The tool returns query results as row objects. To prevent retrieved data from exceeding context window limits, we implement a content-aware truncation strategy: for regular SQL queries, output is restricted to a maximum of 10 rows and overlong cell values (>150 characters) are truncated; for schema-related queries (e.g., \texttt{sqlite\_master} or \texttt{PRAGMA}), the tool preserves the full result set to ensure the agent maintains accurate and complete knowledge of the database structure.

\textbf{Full Text Search.} Databases may contain hundreds of tables and millions of rows, making embedding-based semantic matching expensive to deploy. We instead leverage the SQLite FTS5 extension to construct inverted indices over schema names and database contents, facilitating low-latency fuzzy matching. The tool returns a structured context containing the matched schema elements and up to five distinct matched values per column.

\textbf{Database Profiler.} For each table and column, we prompt an LLM offline to generate a description summarizing semantic intent and to infer foreign-key relationships that may not be formally defined in the DDL. The profiler also computes column-level statistics including distinct value counts, data types, and null-value ratios. By precomputing these static features applicable to all queries, the tool offloads complex schema analysis from the runtime reasoning process, so the agent can focus on query logic rather than schema interpretation.

\begin{figure*}[htb]
    \centering
    \resizebox{0.9\textwidth}{!}{%
      \setlength{\unitlength}{1bp}%
      \begin{picture}(1372,669)
        \put(32,0){\includegraphics[width=1340bp,trim=23.04bp 0 0 0,clip]{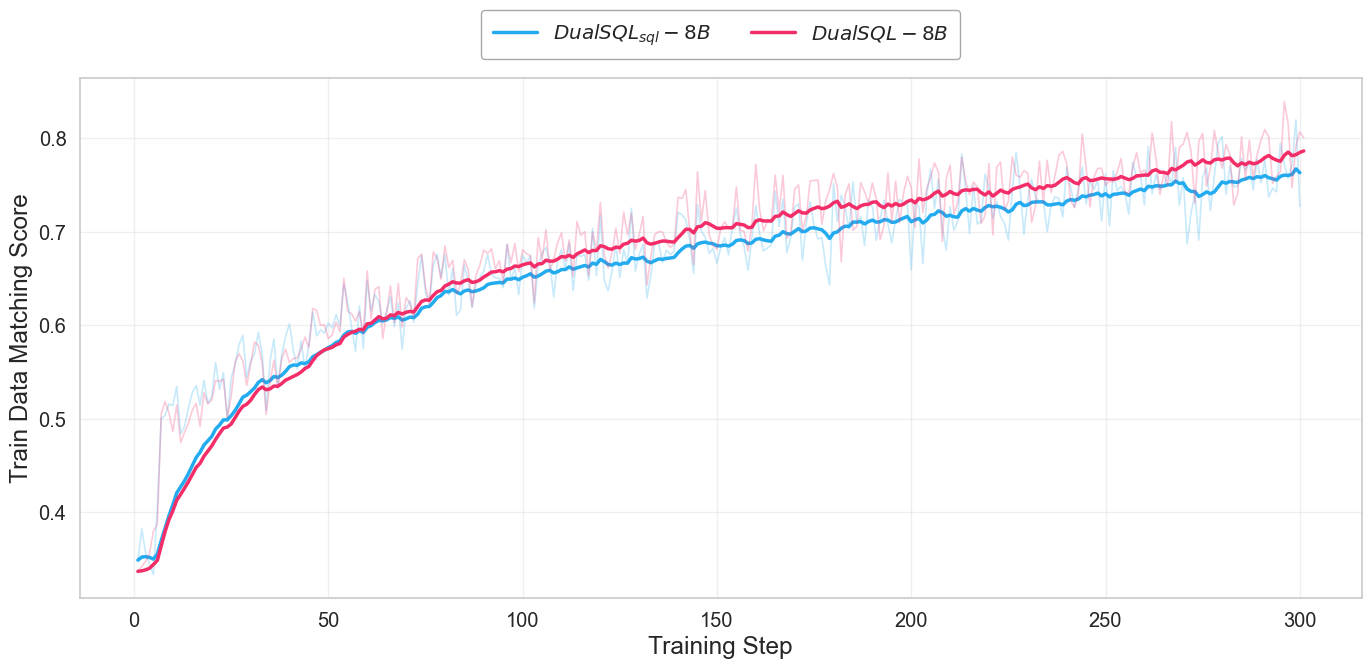}}
        \put(19,331){\makebox(0,0){\rotatebox{90}{\fontfamily{phv}\fontsize{24}{28}\selectfont Train Robust Execution Match (REX)}}}
      \end{picture}%
    }
    \caption{Training robust execution match of $\sysname_{sql}$-8B (single-agent) and \sysname-8B (multi-agent).}
    \label{fig:train_rex_comparison}
\end{figure*}

\section{More Training Details}
\label{appendix:training_details}

We train \sysname-4B for 100 steps. The training for \sysname-8B takes 300 steps, taking about 4.5 days on a 4-node compute cluster, with 8 H100 80GB GPUs per node.

We compare the Robust Execution Match during training for the single-agent $\sysname_{sql}$-8B and multi-agent $\sysname$-8B in Fig.~\ref{fig:train_rex_comparison}. Although $\sysname$-8B starts a little worse due to limited initial schema linking accuracy, it catches up and consistently outperforms $\sysname_{sql}$-8B in the rest of the training process, demonstrating the effectiveness of our multi-agent framework.

\section{Performance Breakdown by Difficulty}
\label{appendix:difficulty_breakdown}

\begin{table*}[htb]
  \setlength{\tabcolsep}{2pt}
  \small
  \centering
  \begin{tabular}{ccccccc}
  \toprule
  Method & Base Model & Task & Easy & Moderate & Challenging  & Overall\\
  \midrule
  SQL-R1-7B & Qwen2.5-Coder-7B & SQL Generation & 72.1 & 60.8  & 51.0  & 66.6 \\
  SQL-R1-14B & Qwen2.5-Coder-14B & SQL Generation & 72.4 & 59.7  & 56.5  & 67.1 \\
  OmniSQL-7B & Qwen2.5-Coder-7B & SQL Generation & - & - & -  &  63.9 \\
  Arctic-text2sql-R1$^*$ & OmniSQL-7B & SQL Generation & 73.9 & 62.3  &55.9 &68.7  \\
  \midrule
  Qwen3-4B  &  - & SQL Generation & 67.5 & 49.1 & 42.6 & 59.6  \\
  $\sysname$-4B (Ours) & Qwen3-4B & Linking + SQL Generation & 73.2 & 61.9 & 54.4 & 68.0 \\
  Qwen3-8B  &  - & SQL Generation & 68.5 & 51.8 & 41.1 &  60.9 \\
  $\sysname_{sql}$-8B (Ours) & Qwen3-8B & SQL Generation & 74.7 & 63.9 & 55.7 & 69.6 \\
  $\sysname$-8B (Ours) & Qwen3-8B & Linking + SQL Generation & \textbf{76.1} & \textbf{64.9} & \textbf{58.7} & \textbf{71.1} \\
  \bottomrule
  \end{tabular}
  \caption{\label{tab:nl2sql_difficulty} Performance on BIRD-Dev by question difficulty. We report average execution match. $^*$We evaluate the released checkpoints of Arctic-text2sql-R1 7B for per-difficulty performance. SQL-R1 and OmniSQL performance are taken from their publications.}
\end{table*}

We break down the performance of $\sysname$ by difficulty in Table \ref{tab:nl2sql_difficulty}. Compared with $\sysname_{sql}$, which is trained only on the SQL generation task with the same recipe, our multi-agent formulation brings performance improvements at all difficulty levels, with the most significant improvement on challenging questions (3.0\% EX for the 8B model).

\begin{table*}[htb]
  \small
  \centering
  \setlength{\tabcolsep}{5.5pt}
\begin{tabular}{lccccccccc}
\toprule
Model & Stage &SQL Execution & Full Text Search & Database Profiler&Total & F1 & EX & REX \\
\midrule
\sysname-8B & Schema Linking & 0.25 & 0.00 & 1.00 & 1.26 & \textbf{90.8} & - & -\\
\sysname-8B & SQL Generation & 1.22 & 0.05 & 0.49 & 1.76 & - & \textbf{71.1} & \textbf{68.6}\\
\midrule
\sysname-8B & Schema Linking & $\times$ & 0.06 & 1.03 & 1.09 & 90.6 & - & -\\
\sysname-8B & SQL Generation & $\times$ & 0.29 & 1.08 & 1.37 & - & 68.6 & 61.8 \\
\sysname-8B & Schema Linking & 0.34 & $\times$ & 1.02 & 1.36 & 90.6 & - & -\\
\sysname-8B & SQL Generation & 1.35 & $\times$ & 0.76 & 2.11 & - & 69.8 & 66.6 \\
\sysname-8B & Schema Linking & 1.24 & 0.15 & $\times$ & 1.39 & 90.2 & - & - \\
\sysname-8B & SQL Generation & 1.43 & 0.27 & $\times$ & 1.70 & - & 69.9 & 66.6 \\
\bottomrule
\end{tabular}
\caption{\label{tab:appendix_ablation_tools}Average number of tools used on BIRD-Dev and overall performance.}
\end{table*}
\section{Ablation on Tools}
We conduct an ablation study to evaluate the impact of each tool on the performance of \sysname-8B on BIRD-Dev.
On the BIRD-Dev dataset, we evaluate the \sysname-8B model, which is trained with all three tools, and disable each tool one at a time. 
We report the average number of tool calls and overall performance in Table~\ref{tab:appendix_ablation_tools}.

\label{appendix:ablation_tools}

The results demonstrate that each tool contributes positively to the overall performance.
Although \sysname~ is able to compensate for the missing tool by increasing the usage of other tools, the removal of any tool results in a performance drop, especially for SQL generation.
The SQL execution tool has the most significant impact when removed.
Without the ability to execute SQL queries and iteratively refine its solutions, the agents suffer a 2.5\% drop in EX and a massive 6.8\% drop in REX during the SQL generation stage.

\newpage
\section{Robust Execution Match}
\label{appendix:data_matching_algorithm}
This algorithm computes a similarity score between a ground truth dataframe ($G$) and a sample dataframe ($S$). Equivalence is determined by $score=1.0$.
The behavior is primarily controlled by a user-defined boolean, \texttt{strict\_row\_ordering}. If columns do not match by name, then they are matched by content using a standard Linear Sum Assignment Problem solver (LSAP) as a subroutine. If the sample dataframe contains more columns/rows than the ground truth, then a proportional penalty is applied. Type differences are handled flexibly.

\begin{algorithm*}
\caption{Robust Execution Match}
\begin{algorithmic}[1]
\REQUIRE $G$, $S$; parameter \texttt{strict\_row\_ordering}
\IF{NOT \texttt{strict\_row\_ordering} AND $\max(|G_{cols}|, |S_{cols}|) > 4$}
  \STATE \hfill // Heuristic for wide dataframes
  \STATE Select $K_{key}$ columns from $G$ with highest cardinality
         (1 or 2 columns depending on $|S_{cols}|$)
  \STATE Align $G[K_{key}]$ and $S$ using exact matching (line 14), yielding row map $\pi_{rows}$
  \STATE Reorder $G$ and $S$ based on $\pi_{rows}$
  \STATE \texttt{strict\_row\_ordering} $\leftarrow$ True \hfill // proceed to strict row logic
  \STATE \textbf{go to line 10}
\ENDIF

\STATE Identify common columns $K_{common}$ and unmatched $K_{u_g}$ (in $G$), $K_{u_s}$ (in $S$)

\IF{\texttt{strict\_row\_ordering}}
  \STATE Score similarity of $G[K_{common}]$ and $S[K_{common}]$ via element-wise comparison
  \STATE Match $K_{u_g}, K_{u_s}$ using LSAP to maximize column similarity, yielding $\pi_{cols}$
  \STATE Combine scores from common and best-matched unmatched columns
\ELSE
  \STATE \hfill // Loose row ordering, exact matching
  \IF{$K_{common} \neq \emptyset$}
    \STATE Match rows of $G[K_{common}]$ and $S[K_{common}]$ to find permutation $\pi_{rows}$ using LSAP
    \STATE Calculate score for $K_{common}$ based on this alignment
    \STATE Reorder $G[K_{u_g}]$ and $S[K_{u_s}]$ according to $\pi_{rows}$
    \STATE Match reordered $K_{u_g}, K_{u_s}$ using column LSAP, yielding $\pi_{cols}$
    \STATE Combine scores
  \ELSE
    \STATE \hfill // No common columns
    \STATE \textbf{Fallback:} iterate column permutations of $S$ with $G$
    \STATE For each permutation, find optimal row alignment $\pi_{rows}$ using LSAP
    \STATE Keep permutation yielding maximum total similarity
  \ENDIF
\ENDIF

\STATE Calculate final score, normalized by $|G_{cols}|$
\STATE Apply penalty: score $\leftarrow$ score $\times \frac{|G_{cols}|}{\max(|G_{cols}|, |S_{cols}|)}$
\STATE Apply penalty: score $\leftarrow$ score $\times \frac{|G_{rows}|}{\max(|G_{rows}|, |S_{rows}|)}$
\STATE \textbf{return} score
\end{algorithmic}
\end{algorithm*}
\subsection*{Complexity Analysis Notes}
Let $N$ be the number of rows and $M$ be the number of columns.

\subsubsection*{Strict Row Ordering}
\begin{itemize}
    \item Element-wise scores: $O(N \cdot M)$.
    \item Unmatched Column Matching (LSAP): Score matrix $O(M^2 \cdot N)$, LSAP solution $O(M^3)$.
    \item \textbf{Total Complexity:} $O(N \cdot M^2)$.
\end{itemize}

\subsubsection*{Loose Row Ordering}
\begin{itemize}
    \item \textbf{Heuristic ($M > 4$):}
        \begin{itemize}
            \item Row alignment on $K_{key}$ (LSAP): $O(N^2 \cdot |K_{key}| + N^3) \approx O(N^3)$.
            \item Column Matching phase: $O(N \cdot M^2)$.
            \item \textbf{Total:} $O(N^3 + N \cdot M^2)$.
        \end{itemize}
    \item \textbf{Exact ($M \leq 4$, With Common Columns):}
        \begin{itemize}
            \item Row alignment on $K_{common}$ (LSAP): $O(N^2 \cdot M + N^3)$.
            \item Column matching on $K_{u}$ (LSAP): $O(M^2 \cdot N)$.
            \item \textbf{Total:} $O(N^3 + N^2 \cdot M)$.
        \end{itemize}
    \item \textbf{Exact ($M \leq 4$, No Common Columns):}
        \begin{itemize}
            \item Column Permutations $\times$ Row LSAP: $O(M! \cdot (N^2 \cdot M + N^3))$.
            \item \textbf{Total:} $O(M! \cdot (N^3 + N^2 \cdot M))$.
        \end{itemize}
\end{itemize}

\section{Prompts}
\subsection{Schema Linking Agent Prompt}
Figure \ref{fig:sl-sys-prompt} and \ref{fig:sl-user-prompt} show the system and user prompts for the Schema Linking Agent. 

\begin{figure*}[htb]
    \centering
    \begin{promptbox}{System Prompt for the Schema Linking Agent}
You are a senior data scientist. Below, you are provided with a set of database access tools, a database schema and a question from users. Your task is to understand the schema and identify the tables and columns that are useful for constructing a SQL query to answer the question.\\

Recall your expertise as a data science expert and solve the question professionally. Your reasoning should be creative and rigorous. Be sure to make good use of the available tools.\\

Key Requirements:\\
- Prioritize coverage but also be mindful about precision. Your results will be used by downstream processes. Therefore, be as accurate as possible to avoid cascading failure.\\
- You have access to a few tools that can help you explore the schema and data. Use them wisely to your advantage.\\

Database Engine: SQLite
    \end{promptbox}
    \caption{System Prompt for Schema Linking Agent}
    \label{fig:sl-sys-prompt}
\end{figure*}

\begin{figure*}[!h]
    \centering
    \begin{promptbox}{User Prompt for Schema Linking Agent}
Database Schema:
\{schema\}
\\
Question: \{question\}\\
Hints: \{hints\}\\
\# Instructions:
In each assistant turn, follow these steps to ensure effective and accurate problem-solving:\\
\#\# Step 1: Reasoning\\
In each assistant turn, you should start your reasoning with the \texttt{<think>} tag, carefully reason about the problem, and end your reasoning with the \texttt{</think>} tag.\\
\#\# Step 2 Action\\
After your reasoning, briefly summarize your thought process. The summary will be visible to the user, therefore, make it concise and clear. Then you should choose to either call tools (2.a) to assist your reasoning, or conclude the conversation with your identified tables and columns (2.b). Show and explain your decision clearly in your response.\\\\
\#\# Step 2.a: Call a tool
You can choose to call any of the provided tools. The execution result of each tool call will be returned to you in the next user turn, wrapped within \texttt{<tool\_response>} \texttt{</tool\_response>} tags. Use this information to inform your next steps. \\\\
\#\# Step 2.b: Submit linked schema\\
When you are confident with your solution, end the conversation with a brief summary of your entire thought process and validation effort, then wrap your solution in a JSON code block. The identified tables should be the key, and the columns should be the values.\\
e.g. \\
\texttt{
```json\\
\{"table1":["column1", "column2"],"table2":["column1"]\}\\
```
}
\\
Keep in mind that you have up to five turns to figure out the solution. Let's work this out turn-by-turn and find the correct schema to use.
    \end{promptbox}
    \caption{User Prompt Template for Schema Linking Agent}
    \label{fig:sl-user-prompt}
\end{figure*}

\subsection{SQL Generation Agent Prompt}
Figure \ref{fig:sql-sys-prompt} and \ref{fig:sql-user-prompt} show the system and user prompts for the SQL Generation Agent.  

\begin{figure*}[htb]
    \centering
    \begin{promptbox}{System Prompt for the SQL Generation Agent}
You are a senior data scientist. Below, you are provided with a set of database access tools, a database schema and a question from users. Your task is to understand the schema and compose a SQL query to answer the question with the help of the tools.\\\\
Recall your expertise as a data science expert and solve the question professionally. Your reasoning should be creative and rigorous. Be sure to make good use of the available tools.\\\\
Key Requirements:\\
- You must validate the correctness of your SQL solution by executing it at least once using the \texttt{`execute\_sql`} tool and checking whether the execution result matches your expectation. Missing this step renders your solution invalid.\\
- Make sure your SQL solution only returns the information asked in the question. If the question asks for a specific column, make sure to only include that column in the SELECT clause, nothing more.\\
- The final SQL query should return all of the information asked in the question without any missing or extra information.\\
- Before generating the final SQL query, please think through the steps of how to write the query.\\
- The tools are idempotent. You do not need to call the same tool with the same arguments more than once. Repeated calls are not encouraged.\\\\

Database Engine: SQLite
    \end{promptbox}
    \caption{System Prompt for SQL Generation Agent}
    \label{fig:sql-sys-prompt}
\end{figure*}

\begin{figure*}[htb]
    \centering
    \begin{promptbox}{User Prompt for the SQL Generation Agent}
Database Schema:
{schema}

Question: {question}
Hints: {hints}

\# Instructions:\\
In each assistant turn, follow these steps to ensure effective and accurate problem-solving:\\

\#\# Step 1: Reasoning\\
In each assistant turn, you should start your reasoning with the \texttt{<think>} tag, carefully reason about the problem, and end your reasoning with the \texttt{</think>} tag. \\

\#\# Reasoning Guidelines:\\
- Avoid hardcoded values in your SQL query as much as possible. Instead, compose useful auxiliary queries into your final solution.\\
- Be especially careful for absent values (NULLs) in the databases. Its impact can be subtle and indirect. But that does not mean you need to always handle NULLs explicitly in your SQL query.\\
- Importantly, you have the privilege to use tools to your advantage. Think carefully how the available tools can help you in your reasoning. When in doubt, use the tools to verify.\\
        
\#\# Step 2 Action\\
After your reasoning, briefly summarize your thought process. The summary will be visible to the user, therefore, make it concise and clear. Then you should choose to either call tools (2.a) to assist your reasoning, or conclude the conversation with your SQL solution (2.b). Show and explain your decision clearly in your response.\\

\#\# Step 2.a: Call a tool\\
You can choose to call any of the provided tools. The execution result of each tool call will be returned to you in the next user turn, wrapped within \texttt{<tool\_response>} \texttt{</tool\_response>} tags. Use this information to inform your next steps. \\

\#\# Step 2.b: Submit SQL solution\\
When you are confident with your solution, end the conversation with a brief summary of your entire thought process and validation effort, then wrap your SQL solution in a code block,\\
e.g. \\
\texttt{
```sql\\
SELECT * FROM table1;\\
```}

Keep in mind that you have up to ten turns to figure out the solution. Let's work this out turn-by-turn and find the correct SQL query.
\\
Keep in mind that you have up to five turns to figure out the solution. Let's work this out turn-by-turn and find the correct schema to use.
    \end{promptbox}
    \caption{User Prompt for the SQL Generation Agent}
    \label{fig:sql-user-prompt}
\end{figure*}


\section{Example Agent Trajectories and Reasoning Patterns}
\label{appendix:example_trajs}

In this section, we showcase the reasoning patterns that emerged through the agents' RL training.

\subsection{Schema Linking Agent}
\begin{figure*}[htb]
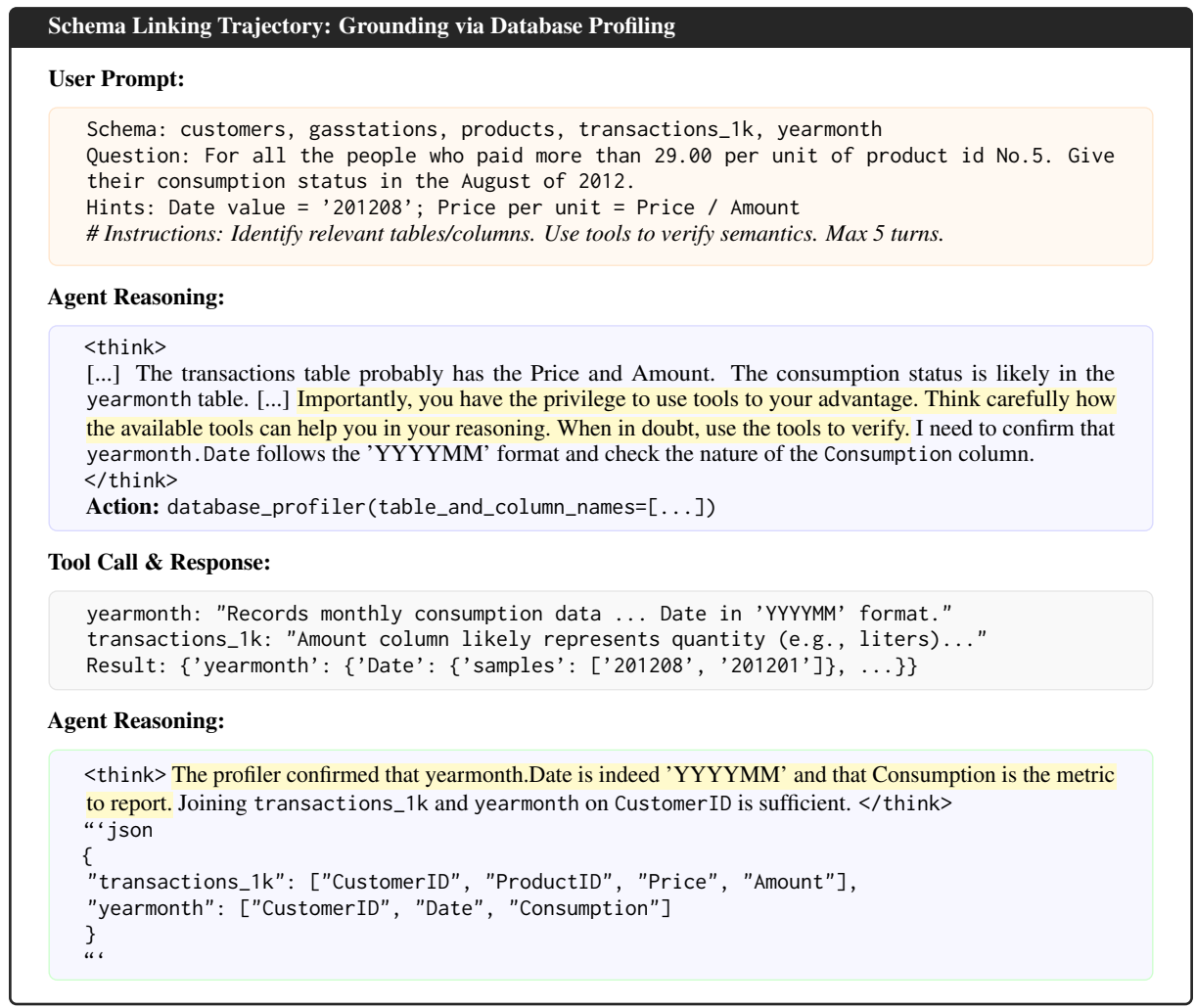

    \centering
    \begin{tcolorbox}[
        title=Schema Linking Trajectory: Grounding via Database Profiling,
        colback=white, 
        colframe=black!85, 
        fonttitle=\bfseries\small,
        arc=2pt,
        boxrule=1pt,
        width=\textwidth,
        enhanced
    ]
        \small
        
        \textbf{User Prompt:}
        \begin{tcolorbox}[colback=orange!5, colframe=orange!20, boxrule=0.5pt, top=2pt, bottom=2pt, enhanced]
            \texttt{Schema: customers, gasstations, products, transactions\_1k, yearmonth}\\{}
            \texttt{Question: For all the people who paid more than 29.00 per unit of product id No.5. Give their consumption status in the August of 2012.}\\{}
            \texttt{Hints: Date value = '201208'; Price per unit = Price / Amount}\\{}
            \vspace{0.3em}
            \textit{\# Instructions: Identify relevant tables/columns. Use tools to verify semantics. Max 5 turns.}
        \end{tcolorbox}

        \vspace{0.3em}

        \textbf{Agent Reasoning:}
        \begin{tcolorbox}[colback=blue!3, colframe=blue!15, boxrule=0.5pt, top=2pt, bottom=2pt, enhanced]
            \texttt{<think>}\\{}
            [...] The transactions table probably has the Price and Amount. The consumption status is likely in the \texttt{yearmonth} table. [...] \hl{Importantly, you have the privilege to use tools to your advantage. Think carefully how the available tools can help you in your reasoning. When in doubt, use the tools to verify.} I need to confirm that \texttt{yearmonth.Date} follows the 'YYYYMM' format and check the nature of the \texttt{Consumption} column.\\{}
            \texttt{</think>}\\{}
            \textbf{Action:} \texttt{database\_profiler(table\_and\_column\_names=[...])}
        \end{tcolorbox}

        \vspace{0.3em}

        \textbf{Tool Call \& Response:}
        \begin{tcolorbox}[colback=gray!5, colframe=gray!20, boxrule=0.5pt, top=2pt, bottom=2pt, enhanced]
            \texttt{yearmonth: "Records monthly consumption data ... Date in 'YYYYMM' format."}\\{}
            \texttt{transactions\_1k: "Amount column likely represents quantity (e.g., liters)..."}\\{}
            \texttt{Result: \{'yearmonth': \{'Date': \{'samples': ['201208', '201201']\}, ...\}\}}
        \end{tcolorbox}

        \vspace{0.3em}

        \textbf{Agent Reasoning:}
        \begin{tcolorbox}[colback=blue!3, colframe=green!20, boxrule=0.5pt, top=2pt, bottom=2pt, enhanced]
            \texttt{<think>} \hl{The profiler confirmed that yearmonth.Date is indeed 'YYYYMM' and that Consumption is the metric to report.} Joining \texttt{transactions\_1k} and \texttt{yearmonth} on \texttt{CustomerID} is sufficient. \texttt{</think>}\\{}
            \texttt{```json}\\
            \texttt{\{} \\
            \texttt{\quad"transactions\_1k": ["CustomerID", "ProductID", "Price", "Amount"],}\\
            \texttt{\quad"yearmonth": ["CustomerID", "Date", "Consumption"]}\\
            \texttt{\}} \\
            \texttt{```}
        \end{tcolorbox}

    \end{tcolorbox}
    \caption{Trajectory illustrating the agent's use of \texttt{database\_profiler} for schema linking. Instead of guessing column formats, it learns to inspect data samples and descriptions, ensuring the final linked schema is sound for the subsequent SQL generation step.}
    \label{fig:schema_linking_profiler}
\end{figure*}

As illustrated in Figure~\ref{fig:schema_linking_profiler}, the Schema Linking agent learns to actively ground its selection process through database profiling rather than relying on zero-shot assumptions. When faced with specific formatting requirements or ambiguous column names, the agent utilizes the \texttt{database\_profiler} tool to inspect actual data samples and column descriptions (e.g., verifying date formats). This ensures the semantic correctness of the chosen tables before proceeding. 

\begin{figure*}[htb]
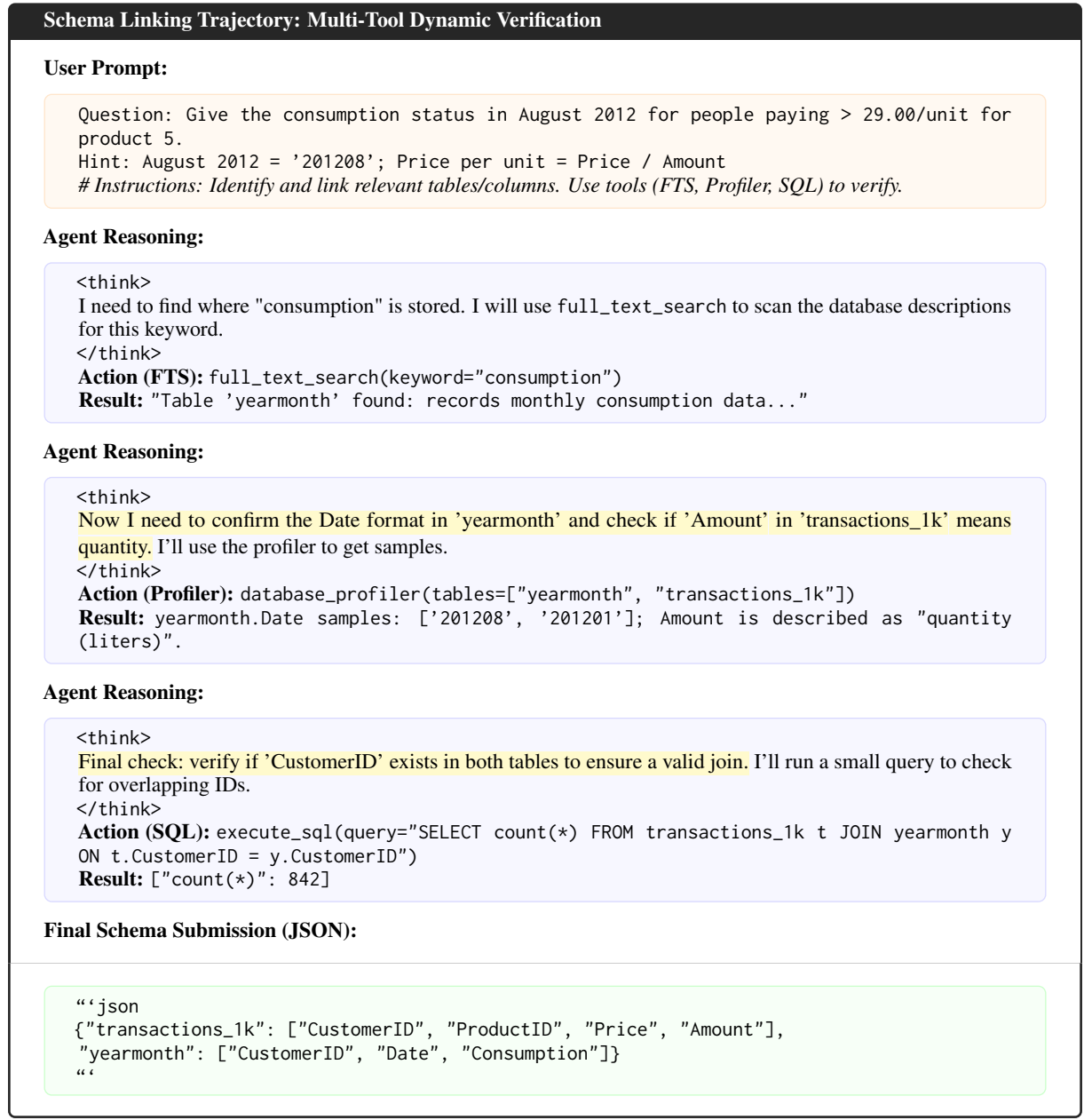

    \centering
    \begin{tcolorbox}[
        title=Schema Linking Trajectory: Multi-Tool Dynamic Verification,
        colback=white, 
        colframe=black!85, 
        fonttitle=\bfseries\small,
        arc=2pt,
        boxrule=1pt,
        width=\textwidth,
        enhanced
    ]
        \small
        
        \textbf{User Prompt:}
        \noindent\begin{tcolorbox}[colback=orange!5, colframe=orange!20,  boxrule=0.5pt, top=2pt, bottom=2pt] 
            \texttt{Question: Give the consumption status in August 2012 for people paying > 29.00/unit for product 5.}\\
            \texttt{Hint: August 2012 = '201208'; Price per unit = Price / Amount}\\
            \textit{\# Instructions: Identify and link relevant tables/columns. Use tools (FTS, Profiler, SQL) to verify.}
        \end{tcolorbox}

        \vspace{0.3em}
        
        \textbf{Agent Reasoning:}
        \noindent\begin{tcolorbox}[colback=blue!3, colframe=blue!15, boxrule=0.5pt, top=2pt, bottom=2pt]
            \texttt{<think>}\\
            I need to find where "consumption" is stored. I will use \texttt{full\_text\_search} to scan the database descriptions for this keyword.\\
            \texttt{</think>}\\
            \textbf{Action (FTS):} \texttt{full\_text\_search(keyword="consumption")}\\
            \textbf{Result:} \texttt{"Table 'yearmonth' found: records monthly consumption data..."}
        \end{tcolorbox}

        \vspace{0.3em}

        \textbf{Agent Reasoning:}
        \noindent\begin{tcolorbox}[colback=blue!3, colframe=blue!15, boxrule=0.5pt, top=2pt, bottom=2pt]
            \texttt{<think>}\\{}
            \hl{Now I need to confirm the Date format in 'yearmonth' and check if 'Amount' in 'transactions\_1k' means quantity.} I'll use the profiler to get samples.\\{}
            \texttt{</think>}\\{}
            \textbf{Action (Profiler):} \texttt{database\_profiler(tables=["yearmonth", "transactions\_1k"])}\\{}
            \textbf{Result:} \texttt{yearmonth.Date samples: ['201208', '201201']; Amount is described as "quantity (liters)".}
        \end{tcolorbox}

        \vspace{0.3em}

        \textbf{Agent Reasoning:}
        \noindent\begin{tcolorbox}[colback=blue!3, colframe=blue!15, boxrule=0.5pt, top=2pt, bottom=2pt]
            \texttt{<think>}\\{}
            \hl{Final check: verify if 'CustomerID' exists in both tables to ensure a valid join.} I'll run a small query to check for overlapping IDs.\\{}
            \texttt{</think>}\\{}
            \textbf{Action (SQL):} \texttt{execute\_sql(query="SELECT count(*) FROM transactions\_1k t JOIN yearmonth y ON t.CustomerID = y.CustomerID")}\\{}
            \textbf{Result:} \texttt{[{"count(*)": 842}]}
        \end{tcolorbox}

        \vspace{0.3em}

        \textbf{Final Schema Submission (JSON):}
        \noindent\begin{tcolorbox}[colback=green!3, colframe=green!20, boxrule=0.5pt, top=2pt, bottom=2pt, enhanced]
            \texttt{```json}\\
            \texttt{\{"transactions\_1k": ["CustomerID", "ProductID", "Price", "Amount"],}\\
            \texttt{ "yearmonth": ["CustomerID", "Date", "Consumption"]\}}\\
            \texttt{```}
        \end{tcolorbox}

    \end{tcolorbox}
    \caption{Schema Linking trajectory for adaptive schema pruning based on tool calling responses from \textbf{Full-Text Search}, \textbf{Database Profiler}, and \textbf{Execute SQL}.}
    \label{fig:multi_tool_schema_linking}
\end{figure*}

Furthermore, Figure~\ref{fig:multi_tool_schema_linking} demonstrates a more sophisticated, multi-tool dynamic verification strategy. In this trajectory, the agent sequentially chains multiple tools to adaptively prune the schema: it employs \texttt{full\_text\_search} to locate relevant tables based on keywords, profiles the data to confirm formats, and executes exploratory queries via \texttt{execute\_sql} to empirically validate table joins (e.g., checking for overlapping entity IDs between tables). Together, these emergent reasoning patterns highlight the agent's ability to iteratively gather evidence and construct a highly reliable schema for downstream SQL generation.

\subsection{SQL Generation Agent}
\begin{figure*}[htb]
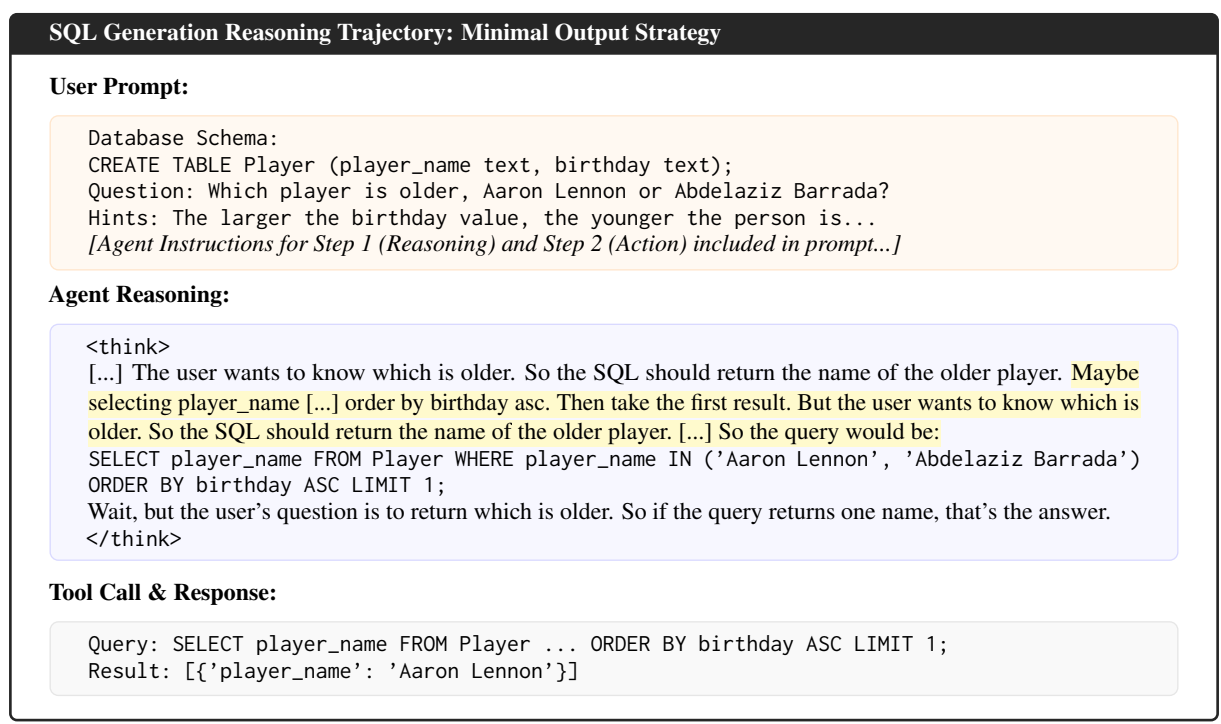

    \centering
    \begin{tcolorbox}[
        title=SQL Generation Reasoning Trajectory: Minimal Output Strategy,
        colback=white, 
        colframe=black!85, 
        fonttitle=\bfseries\small,
        arc=2pt,
        boxrule=1pt,
        width=\textwidth,
        enhanced
    ]
        \small
        
        \textbf{User Prompt:}
        \begin{tcolorbox}[colback=orange!5, colframe=orange!20, boxrule=0.5pt, top=2pt, bottom=2pt, enhanced]
            \texttt{Database Schema:}\\{}
            \texttt{CREATE TABLE Player (player\_name text, birthday text);}\\{}
            \texttt{Question: Which player is older, Aaron Lennon or Abdelaziz Barrada?}\\{}
            \texttt{Hints: The larger the birthday value, the younger the person is...}\\{}
            \textit{[Agent Instructions for Step 1 (Reasoning) and Step 2 (Action) included in prompt...]}
        \end{tcolorbox}
   
        \textbf{Agent Reasoning:}
        \begin{tcolorbox}[colback=blue!3, colframe=blue!15, boxrule=0.5pt, top=2pt, bottom=2pt]
            \texttt{<think>}\\{}
            [...] The user wants to know which is older. So the SQL should return the name of the older player. \hl{Maybe selecting player\_name [...] order by birthday asc. Then take the first result. But the user wants to know which is older. So the SQL should return the name of the older player. [...] So the query would be:}\\{}
            \texttt{SELECT player\_name FROM Player WHERE player\_name IN ('Aaron Lennon', 'Abdelaziz Barrada') ORDER BY birthday ASC LIMIT 1;}\\{}
            Wait, but the user's question is to return which is older. So if the query returns one name, that's the answer.\\{}
            \texttt{</think>}
        \end{tcolorbox}

        \vspace{0.3em}

        \textbf{Tool Call \& Response:}
        \begin{tcolorbox}[colback=gray!5, colframe=gray!20, boxrule=0.5pt, top=2pt, bottom=2pt]
            \texttt{Query: SELECT player\_name FROM Player ... ORDER BY birthday ASC LIMIT 1;}\\{}
            \texttt{Result: [\{'player\_name': 'Aaron Lennon'\}]}
        \end{tcolorbox}

    \end{tcolorbox}
    \caption{SQL Generation trajectory showing the agent's decision-making process to ensure the SQL output is minimal, just answering what the user asked for using \texttt{LIMIT 1}.}
    \label{fig:sql_minimal_output}
\end{figure*}

For the SQL Generation agent, we observe a strong tendency toward a minimal output strategy, as depicted in Figure~\ref{fig:sql_minimal_output}. Rather than generating verbose queries that return extraneous columns or intermediate calculations, the agent carefully aligns its SQL logic with the exact semantic intent of the user's question. For instance, when asked to identify which of two players is older, the agent explicitly reasons that the final output should only be a single name. It subsequently formulates a query utilizing an \texttt{ORDER BY} clause combined with \texttt{LIMIT 1} to extract precisely the requested entity. This behavior indicates that the agent has learned to synthesize highly targeted queries that directly answer the prompt without requiring further human interpretation of the result set.

Furthermore, the agent exhibits advanced iterative debugging capabilities and an awareness of database-specific nuances, as illustrated in Figure~\ref{fig:sql_cast_trajectory}. In this trajectory, the agent's initial query execution unexpectedly yields \texttt{NULL} values. Instead of halting or blindly guessing a fix, the agent systematically hypothesizes potential mathematical and dialect-specific causes—such as division by zero or SQLite's default handling of integer division. By dynamically reasoning about the underlying database engine's mechanics, the agent actively corrects its query by applying the \texttt{CAST(... AS REAL)} function to prevent integer truncation. This demonstrates how the RL training framework enables the agent to utilize execution feedback not just to fix syntax errors, but to resolve subtle, data-dependent semantic edge cases.

\FloatBarrier

\definecolor{highlightyellow}{RGB}{255, 250, 205}
\sethlcolor{highlightyellow}

\begin{figure*}[htb]
    \centering
    \begin{tcolorbox}[
        title=SQL Generation Reasoning Trajectory: Cast to avoid Null,
        colback=white, 
        colframe=black!85, 
        fonttitle=\bfseries\small,
        arc=2pt,
        boxrule=1pt,
        enhanced,
        width=\textwidth
    ]
        \small

        \textbf{User Prompt:}
        \begin{tcolorbox}[colback=orange!5, colframe=orange!20, boxrule=0.5pt, top=2pt, bottom=2pt]
            \texttt{Database Schema:}\\{}
            \texttt{CREATE TABLE frpm (}\\{}
            \texttt{\quad `Educational Option Type` text, -- ['Continuation School', ...]}\\{}
            \texttt{\quad `Enrollment (Ages 5-17)` real,}\\{}
            \texttt{\quad `Free Meal Count (Ages 5-17)` real );}\\{}
            \vspace{0.3em}
            \texttt{Question: Please list the lowest three eligible free rates for students aged 5-17 in continuation schools.}\\{}
            \texttt{Hints: Rate = `Free Meal Count` / `Enrollment`}\\{}
            \vspace{0.3em}
            \textit{\# Instructions: Follow Step 1 (Reasoning in <think> tags) and Step 2 (Action: Tool call or Submit SQL). Max 10 turns.}
        \end{tcolorbox}

        \vspace{0.3em}

        \textbf{Agent Reasoning:}
        \begin{tcolorbox}[colback=blue!3, colframe=blue!15, boxrule=0.5pt, top=2pt, bottom=2pt]
            \texttt{<think>}\\{}
            [...] Maybe the problem is that some rows have zero enrollment, leading to division by zero. Let's check if enrollment is zero. [...] Let me adjust the query to add a \texttt{WHERE} clause for enrollment not being zero.\\{}
            \texttt{</think>}
        \end{tcolorbox}

        \vspace{0.3em}
        
        \textbf{Tool Call \& Response:}
        \begin{tcolorbox}[colback=gray!5, colframe=gray!20, boxrule=0.5pt, top=2pt, bottom=2pt]
            \texttt{Query: SELECT `Free Meal Count` / `Enrollment` AS rate FROM ...}\\{}
            \texttt{Result: [\{'eligible\_rate': None\}, \{'eligible\_rate': None\}]}
        \end{tcolorbox}

        \vspace{0.3em}

        \textbf{Agent Reasoning:}
        \begin{tcolorbox}[colback=blue!3, colframe=blue!15, boxrule=0.5pt, top=2pt, bottom=2pt]
            \texttt{<think>}\\{}
            Hmm, the result is still all NULLs. That's not right. [...] Wait, maybe the division is not being computed correctly. \hl{Wait, in SQLite, if the division is between integers, it would do integer division. But the schema says they are real. Maybe the data is stored as integers, but the division is not working as expected. So maybe the correct way is to cast them as real.}\\ {}
            So, perhaps the query should be: \texttt{SELECT CAST(count AS REAL) / CAST(enrollment AS REAL)...} Let me check the actual values for these rows first.\\{}
            \texttt{</think>}
        \end{tcolorbox}
        
    \end{tcolorbox} 
    \caption{SQL Generation trajectory showing the agent identifying an integer division issue in SQLite through iterative tool use and reasoning.}
    \label{fig:sql_cast_trajectory}
\end{figure*}


\end{document}